\documentclass[letterpaper, 10 pt, conference]{ieeeconf}  

\usepackage{graphicx} 
\usepackage{float} 
\usepackage{subfigure} 
\usepackage{cite}
\usepackage{color}
\usepackage{url}
\usepackage{array}
\usepackage{booktabs} 
\usepackage{multirow} 
\usepackage{makecell}
\usepackage{threeparttable}
\usepackage{hyperref}

\IEEEoverridecommandlockouts                              

\title{\LARGE \bf
CRRS: Concentric Rectangles Regression Strategy for Multi-point Representation on Fisheye Images
}

\author{Xihan Wang$^{1,2}$, Xi Xu$^{1,2}$, Yu Gao$^{1,2}$, Yi Yang$^{*,1,2}$, Yufeng Yue$^{1,2}$, Mengyin Fu$^{1,2}$
\thanks{$^{1}$School of Automation, Beijing Institute of Technology, Beijing, China}%
\thanks{$^{2}$State Key Laboratory of Intelligent Control and Decision of Complex
System, Beijing Institute of Technology, Beijing, China}%
\thanks{*Corresponding author: Y. Yang Email: yang\_yi@bit.edu.cn}%
}

\begin{document}

\maketitle
\thispagestyle{empty}
\pagestyle{empty}


\begin{abstract}

Modern object detectors take advantage of rectangular bounding boxes as a conventional way to represent objects. When it comes to fisheye images, rectangular boxes involve more background noise rather than semantic information. Although multi-point representation has been proposed, both the regression accuracy and convergence still perform inferior to the widely used rectangular boxes. In order to further exploit the advantages of multi-point representation for distorted images, Concentric Rectangles Regression Strategy(CRRS) is proposed in this work. We adopt smoother mean loss to allocate weights and discuss the effect of hyper-parameter to prediction results. Moreover, an accurate pixel-level method is designed to obtain irregular IoU for estimating detector performance. Compared with the previous work for muti-point representation, the experiments show that CRRS can improve the training performance both in accurate and stability. We also prove that multi-task weighting strategy facilitates regression process in this design. Source code is at \textcolor[rgb]{0,0,1}{\url{https://github.com/IN2-ViAUn/Concentric-Rectangles-Loss}}.
\end{abstract}

\section{INTRODUCTION}

\begin{figure}[htpb]
\
\includegraphics[width=0.48\textwidth]{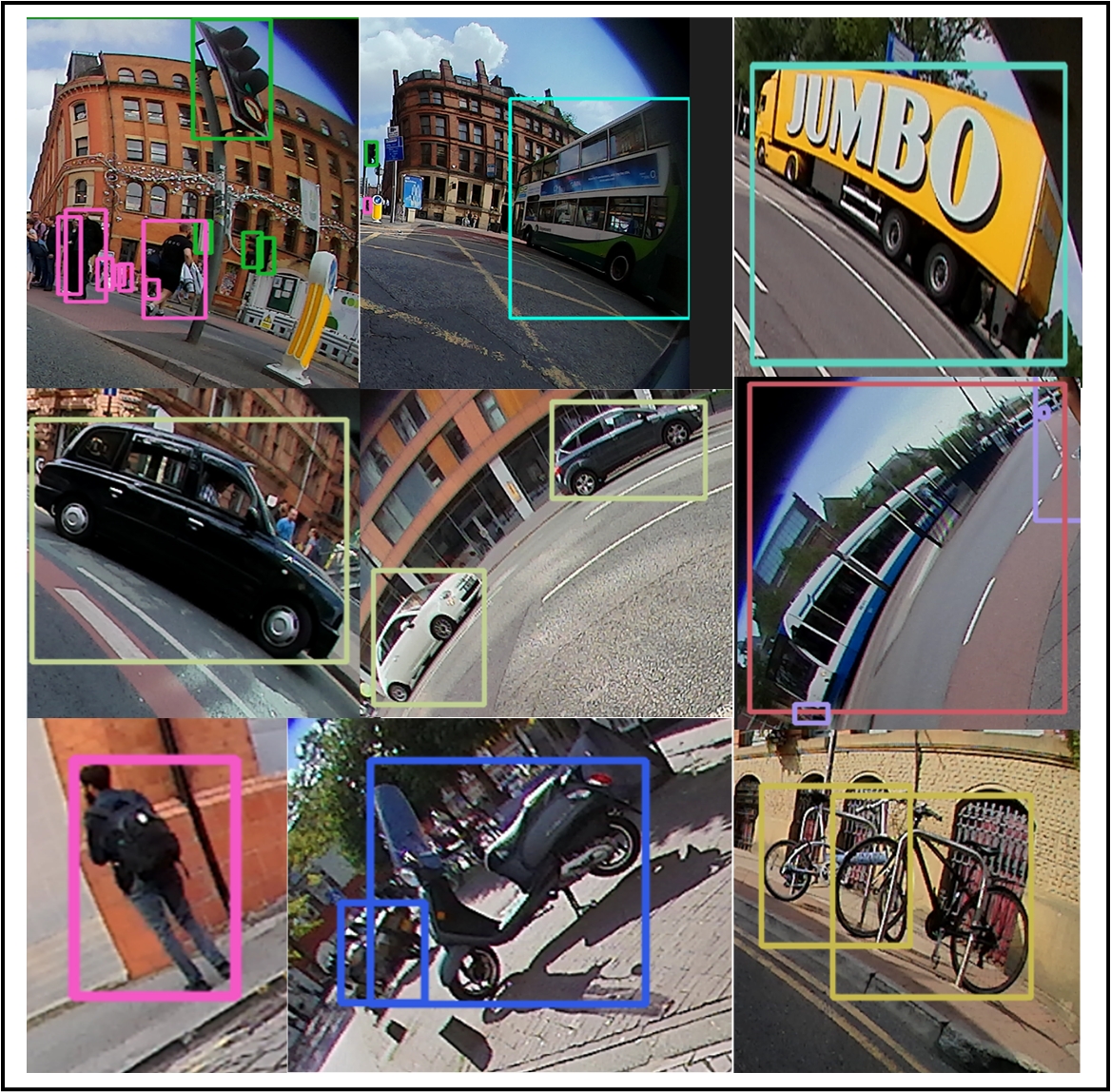}
\caption{Rectangular bounding boxes provide rough representations for most objects in distorted fisheye images.}
\label{box vis}
\end{figure}

The representation of standard rectangular bounding boxes is widely applied in object detectors. Despite being more convenient to calculate coordination, rectangular boxes include much invalid background information which interferes object detection, as shown in Fig.\ref{box vis}. 
This drawback becomes more obvious for fisheye images due to severe distortion. Therefore, it is significant to explore novel representations for fisheye images. To solve this problem, relevant researches have proposed modified boxes, like rotation boxes\cite{index-2,index-3}, circles\cite{index-5} and ellipses\cite{index-4}. Particularly, FisheyeDet\cite{index-7} utilized irregular quadrangle to represent distorted objects. Multi-point representation, proposed by Rashed et.al\cite{index-8,index-9}, can exactly describe the outline of objects.

\par Multi-point representation faces two challenges: (1) The multi-point regression strategy encounters complicated calculation and convergence difficulty. When each point is regressed separately, the number of loss items increase conspicuously. This phenomenon may lead to slow convergence or even failure\cite{index-10}. Therefore, multi-point loss function requires considering both efficiency and convergence speed.  (2) Compared with rectangular boxes, computation complexity of multi-point Intersection over Union(IoU) increases dramatically. Specifically, there are many stages requiring IoU calculation, such as label assignment, loss function, and mean Average Precision(mAP). The more time-consuming IoU calculates, the more time costs, which is necessary to be explored for more efficient means.

\begin{figure*}[htpb]
\centering
\includegraphics[width=\textwidth]{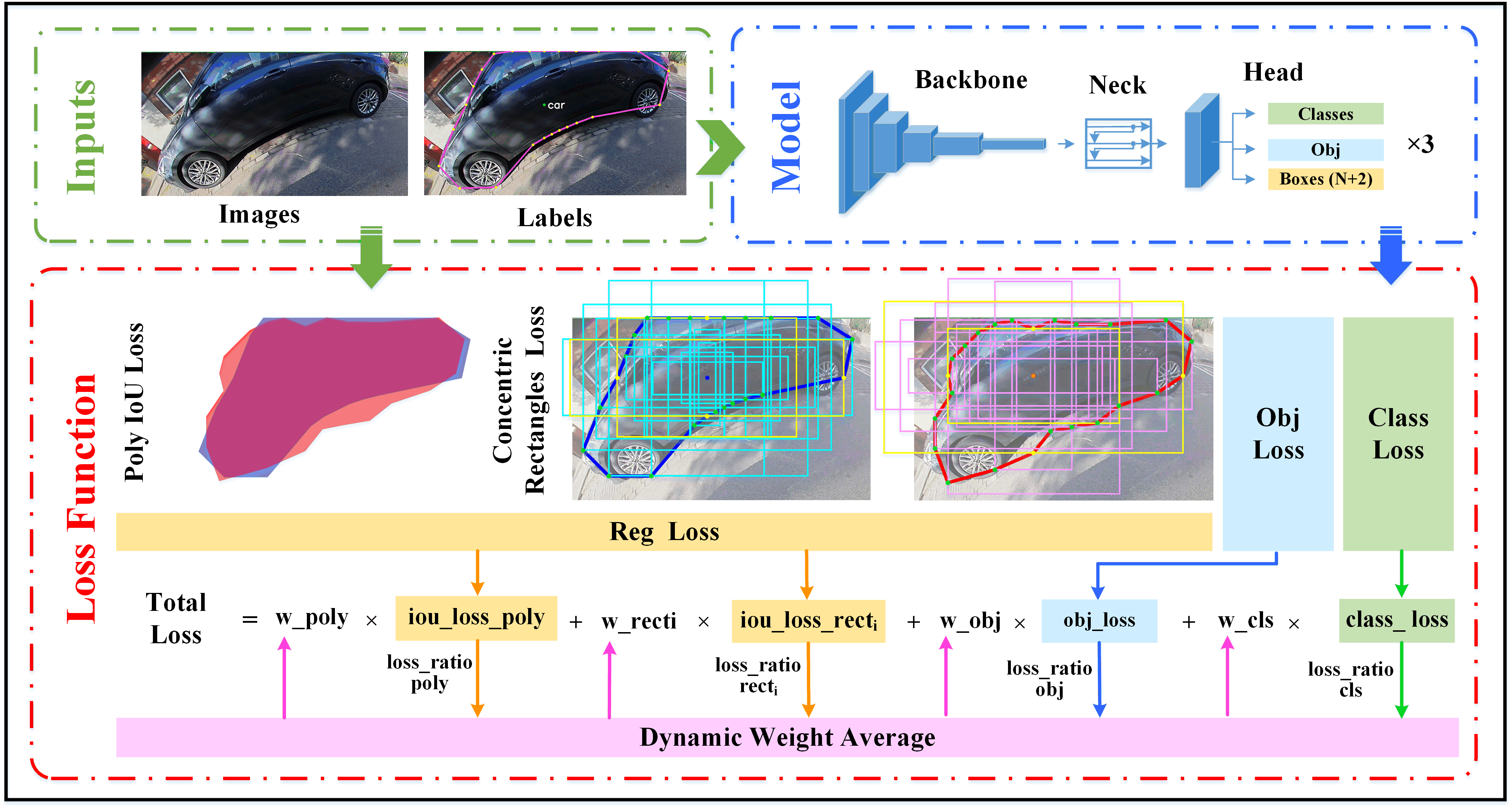}
\caption{\textbf{{Overview of the experiment process}}. Inputs: distorted images and multi-point labels. Model: the outputs of network is changed for multi-point with N distances and the centroid coordinates(Fig.\ref{22rect}). Loss function: Reg Loss is the key contribution which composed of Poly IoU loss and Concentric Rectangles Loss. Dynamic Weights Average(DWA) is used to allocate weights. }
\label{overview}
\end{figure*}

\par Aimed at the above problems, the following solutions are proposed: (1) A regression strategy called Concentric Rectangles Loss is elegantly designed to promote the prediction performance. Firstly, we design a series of concentric rectangles for regression. The reason why using rectangles is that circle shape is merely decided by radius, causing advanced IoU loss unavailable. We further analyze the necessity of concentric feature based on previous work\cite{index-11}. Secondly, we introduce weighting strategy to accelerate convergence. Because Concentric Rectangles Loss leads to a large number of loss items, adaptive loss weight strategies are needed to balance the convergence rates of each loss. Dynamic Weight Average(DWA)\cite{index-12} is verified feasible to solve this problem. Furthermore, we discuss the quality of weight allocation under different temperature coefficients. Also, calculating loss ratio by mean loss is able to reduce the negative impact of single loss fluctuations.
(2) A pixel-level method for polygon IoU is elegantly designed to generate accurate results. The relationship of pixel and polygon is altered into that of pixel and triangle. This approach can be applied to calculate polygon area composed of different numbers of points and diverse sampling ways. Polygon IoU plays an important role in evaluating detection performance. Precise mAP provides quantitative comparison between different regression strategies for multi-point representation in object detection. 
\par The main contributions of this paper are as follows:

\par $\bullet$ CRRS is proposed for multi-point regression. The multi-task learning(MTL) weighting strategies are adopted in this paper to improve convergence speed. Furthermore, we propose the efficient combination of multiple loss. 
\par $\bullet$ We design a way to calculate IoU for multi-point representation in object detection task. On the basis of Polygon IoU, it is feasible to evaluate prediction performance quantitatively.

\section{RELATED WORK}

\subsection{Object Representation}

\par Rectangular boundary box with regular shape and simple computation is convenient for position regression. Currently, it is one of the most popular representations of object detector\cite{index-13,index-14,index-16}. However, rectangular boxes always contain much background information. This feature exerts a negative influence on the learning of semantic information and object detection model. Unfortunately, this defect is more serious when the images are distorted or the objects are dense and severely hidden. Therefore, more suitable representation methods are proposed accord with object shape and position. It is common to use oriented boxes to conduct people detection in overhead fish eye images\cite{index-17,index-18}. Compared with rectangular box, circle box\cite{index-20,index-21} only requires radius prediction with one Degree of Freedom(DoF) and is rotation invariant. Ellipse box\cite{index-4} can include less background area, so it is more adaptive for severely overlapped objects. In addition, lots of researches aim to represent distorted objects more exactly with reduced background information. Thus, multi-point representation, like irregular quadrilateral box\cite{index-7} and 24-side box\cite{index-9,index-11} is put forward.These representation methods are able to extract feature to locate objects more precisely.\par IoU matters in label allocation, loss function, evaluation index and non-maximum suppression(NMS). Various representations make irregular IoU calculation more difficult. It is relatively simple to compute IoU of two circles\cite{index-21}, but the condition of two polygon is more complicated. Li et.al\cite{index-7}took advantage of cross product of vector. Based on statistic approach, they proposed a method to estimate the positional relationship of sampling point and irregular quadrangle. Xie et.al\cite{index-10}put forward Polar IoU, an algorithm to calculate mask IoU on the basis of polar vector. In this way, the complex IoU computation of polygon prediction box and truth box can be avoided. Previous work\cite{index-11} adopted the mean of 24 concentric circles IoU in the stage of label allocation as approximate value. But on account of the lack of accurate calculation of 24 polygon IoU, there is no way of describing prediction performance quantitatively by mAP. For the distorted objects in 360°images, Cao et.al\cite{index-22} proposed a new detection method using Field-of-View Bounding Boxes to replace rectangular boxes. Also, they proposed an approximate method called FoV-IoU which is more accurate than Sph-IoU\cite{index-23}. This method helps object detectors to achieve better prediction results during training, inference, and evaluation stages. For multi-poins representation, this paper proposes a pixel-level method to accurately calculate polygon IoU and obtain accurate mAP evaluation results.
\begin{figure*}[htpb]
\centering
\includegraphics[width=\textwidth]{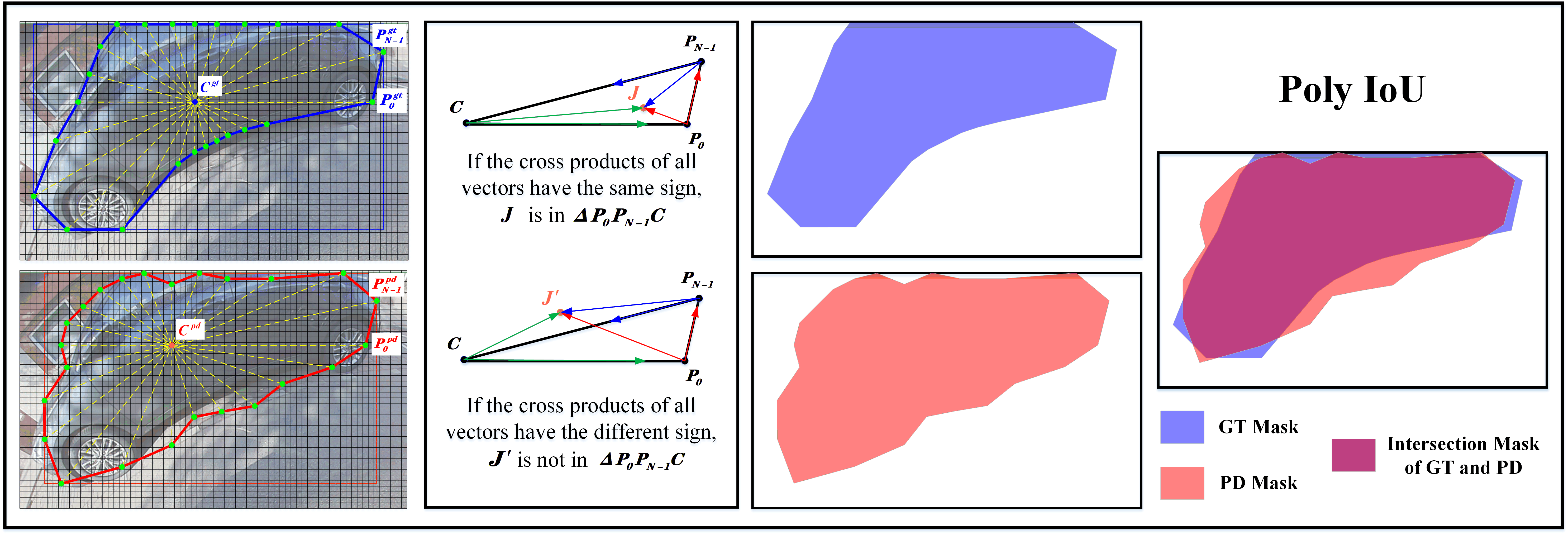}
\caption{The procedure of calculating Poly IoU.}
\label{poly iou}
\end{figure*}
\subsection{Weighting Strategies of Multi-task Loss}

MTL\cite{index-24} is widely used in the field of computer vision, like classification, depth estimation and semantic dense prediction. Since different tasks have different objective functions, it is necessary to consider how to allocate task loss weights when training a multi-task model. Therefore, many researchers have proposed various multi-task loss weight adjustment strategies. Different tasks can extract features by hard parameter sharing or soft parameter sharing for the best learning performance. Because diverse tasks may be unbalanced during training process, setting fixed weight for every task is no longer available. It is necessary to set adaptive weights in the loss function through unique strategies. The aim that all the tasks are synchronous\cite{index-25} or  auxiliary task accelerates main task\cite{index-26,index-27} is achieved thanks to these weighting methods. \par Bayesian uncertainty can be used to design weight parameter\cite{index-29,index-30}. This method weights the losses of different tasks based on the uncertainty of the task loss. Specifically, the uncertainty weighting strategy weights the loss function of different tasks to reflect the difficulty of the task. Later some researches utilized learning rate to assign weighting for different losses\cite{index-12,index-31,index-32}. Also there are certain classical algorithms update weight by gradient\cite{index-33,index-34}.The gradient sharing strategy is a method that combines the loss functions of multiple tasks into a total loss function. This strategy allows different tasks to share model parameters to increase training speed and reduce the risk of over-fitting. Specifically, the gradient sharing strategy can balance more difficult tasks by using lower loss weights. GradNorm\cite{index-33} balances learning among tasks by normalizing the gradient norm to a common scale or considering the relationship between the gradient directions of each task's loss function. HydaLearn\cite{index-35} introduces a dynamic weighted method for multi-task learning loss with auxiliary tasks. However, the above methods are mainly used for weight adjustment among multiple tasks. In addition, the information entropy-based strategy is a method that assigns task weights based on the mutual information between tasks. Specifically, this strategy calculates the mutual information between tasks to assign task weights. If two tasks have high mutual information, their weights will be adjusted to higher weights, and vice versa.If the special problem of single-task multi-loss is also considered as multi-task learning, the weight adjustment strategy of certain methods is no longer applicable. \cite{index-36} has proven that for single-task multi-loss problems, using the coefficient of variation to design weight adjustment strategies can achieve better results. This paper synthesizes DWA\cite{index-12} and Welford algorithm\cite{index-37} to dynamically adjust the weights of single-task multi-loss.

\section{METHOD}

Fig.\ref{overview} summarizes the main research work of this paper. It mainly involves the design of the position regression loss term for multi-point representation and multiple loss weight adjustment strategies. Our research methods will be introduced from three aspects: Polygon IoU Calculation, Concentric Rectangles Loss and Adaptive Weighting Strategies of Multi-loss. In order to compare with previous work\cite{index-11}, this paper still uses a 24-sided polygon to represent the object. $N$ is the number of vertices of the polygon bounding box and equals to 24.

\subsection{Polygon IoU Calculation} \textbf{Polygon bounding box}: For the ground truth box, we first determine $C^{gt}(x_{c}^{gt},y_{c}^{gt})$ as the centroid of each object based on the instance mask. The formula is as follows:

\begin{equation}
x_{c}^{gt}=\frac{M_{10}}{M_{00}},\quad y_{c}^{gt}=\frac{M_{01}}{M_{00}}\label{1}
\end{equation}where the zero-order moment of the object contour $M_{00}=\sum_{i}\sum_{j}V(i,j)$, the first moment is respectively: $M_{10}=\sum_{i}\sum_{j}i \cdot V(i,j)$ and $M_{01}=\sum_{i}\sum_{j}j \cdot V(i,j)$, $i$ and $j$ respectively represent the horizontal and vertical coordinates of each pixel point composing the object contour, $V(i,j)$represents the gray-scale value of $(i,j)$ pixel. For distorted objects, centroid outweighs center of rectangular bounding box. There may even be probable that the center falls outside the object mask, which affects the confirmation of vertices. We obtained $N$ boundary points using equal-angular sampling: $P_{0}^{gt}(x_{0}^{gt},y_{0}^{gt}),P_{1}^{gt}(x_{1}^{gt},y_{1}^{gt}),...,P_{N-1}^{gt}(x_{N-1}^{gt},y_{N-1}^{gt})$. In these points, $P_{0}^{gt}$ is the intersection point between the ray emitted from the centroid $C^{gt}$ in the direction of the positive half-axis of the pixel coordinate system and the object contour. Then every $360^\circ /N$, other points are sampled clockwise to form an approximate polygonal ground truth box for the object contour.\par For the prediction box, we calculate $N$ boundary points $P_{0}^{pd}(x_{0}^{pd},y_{0}^{pd}),P_{1}^{pd}(x_{1}^{pd},y_{1}^{pd}),...,P_{N-1}^{pd}(x_{N-1}^{pd},y_{N-1}^{pd})$ based on the outputs of the network$(x_{c}^{pd},y_{c}^{pd},r_{0},r_{1},...,r_{N-1})$ and Formula (2):

$$
\left\{ {\begin{array}{*{20}{c}}
{x_k^{pd} = x_c^{pd} + {r_k} \cdot \cos (k \cdot \frac{{360^\circ }}{N})}\\
{y_k^{pd} = y_c^{pd} + {r_k} \cdot \sin (k \cdot \frac{{360^\circ }}{N})}
\end{array}} \right. \eqno{(2)}
$$where $k=0,1,...,N-1$, $(x_{c}^{pd},y_{c}^{pd})$ is the centroid coordinates predicted by the network. $r_{0},r_{1},...,r_{N-1}$ are the Euclidean distance between the centroid predicted and $N$ predicted boundary points.\par \textbf{The calculation of irregular IoU}:  The complexity of overlapping shape between two polygons is much greater than that of two rectangles. Therefore, we propose a pixel-level confirmation method that can accurately calculate the area of any polygon and the IoU. The polygon is constructed by the above method can be easily divided into $N$ triangles. Since it is relatively simple to determine whether a point is inside a triangle, the problem of calculating the polygon area can be simplified. It is more easy to confirm whether the pixels contained in the circumscribed rectangle of the polygon are inside any of the triangles. As shown in Fig.\ref{poly iou}, the triangle composed of $P_{0}$, $P_{N-1}$, $C$ and the point $J$ can illustrate the judging method. Construct three sets of vectors $\overrightarrow {{P_0}{P_{N - 1}}} $ and $\overrightarrow {{P_0}J} $, $\overrightarrow {{P_{N - 1}}C} $ and $\overrightarrow {{P_{N - 1}}J} $, $\overrightarrow {C{P_0}} $ and $\overrightarrow {CJ} $. Perform a cross product calculation on each set of vectors: 

$$
\begin{array}{l}
{cross1 = \overrightarrow {{P_0}{P_{N - 1}}}  \times \overrightarrow {{P_0}J}} \\
{cross2 = \overrightarrow {{P_{N - 1}}C}  \times \overrightarrow {{P_{N - 1}}J} }\\
{cross3 = \overrightarrow {C{P_0}}  \times \overrightarrow {CJ} }
\end{array}  \eqno{(3)}
$$

If $cross1$, $cross2$ and $cross3$ have same sign, point $J$ is in the triangle. In order to compute area, we set the gray-scale value of pixels inside the polygon to 1, and set the rest to 0. Summing up the gray-scale values of all pixels will give the area of the polygon. The key of calculating IoU is to compute the intersection area. We first determine which pixels are included in the ground truth polygon and the predicted polygon respectively. After converting the masks of the two polygons to Boolean types, the "and" operation is performed. The number of intersection pixels is counted to calculate the polygon IoU(Poly IoU). Our method has been validated to be applicable to both convex and concave polygons.

\begin{figure*}[htpb]
\centering
\includegraphics[width=\textwidth]{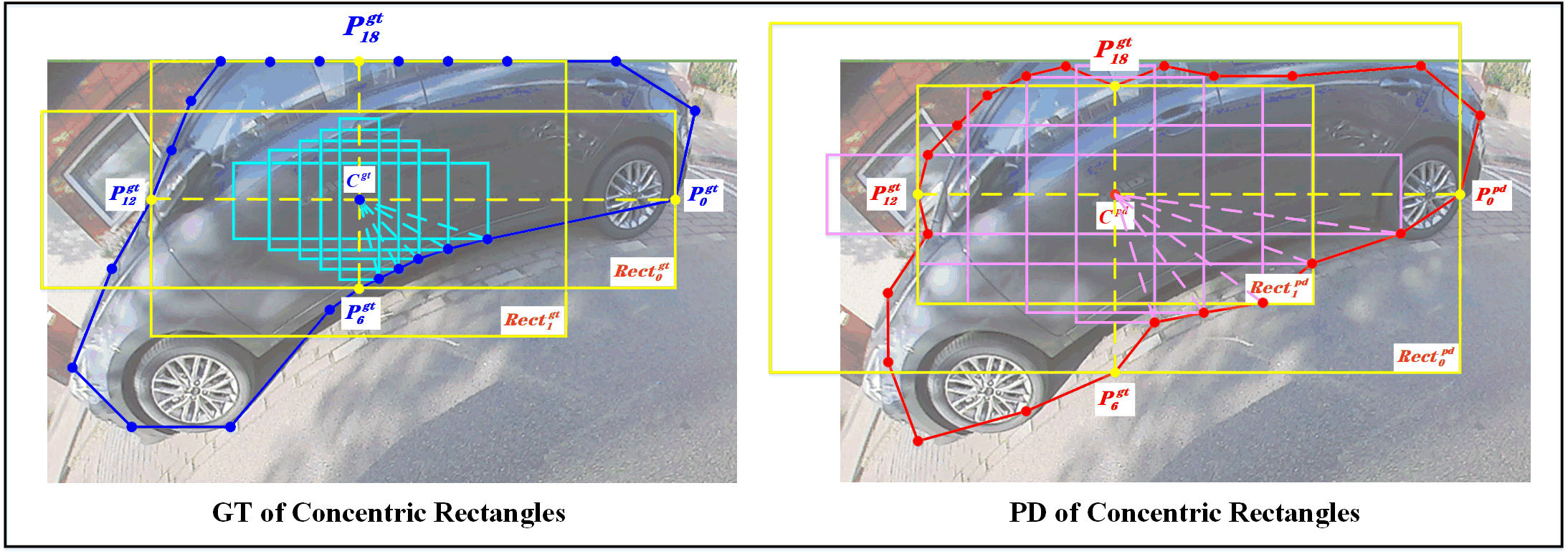}
\caption{Illustration of partial points and centroid for concentric rectangles.}
\label{22rect}
\end{figure*}

\subsection{Concentric Rectangles Loss}
We designed a Concentric Rectangles Regression Strategy with EIoU loss\cite{index-39} inspired by the previous concentric circles(the design of aspect ratio in EIoU loss and CIoU loss\cite{index-38} cannot help with circles regression). This new strategy not only accelerates convergence but also efficiently and accurately completes the polygonal position regression by fully utilizing Eiou Loss.

\textbf{The construction of concentric multi-rectangle:} As shown in Fig.\ref{22rect}, four bounding points $P_{0}^{gt}$,$P_{6}^{gt}$,$P_{12}^{gt}$,$P_{18}^{gt}$ are aligned with the centroid axis. So we combine them to form 
$Rect_{0}^{gt}$ and $Rect_{1}^{gt}$. The remaining boundary points form rectangles with the centroid respectively and the total number of rectangles is $N-1$. Each rectangle has its center point aligned with the centroid. The Euclidean distance from the centroid to the boundary points is taken as half of the diagonal of the rectangle, so we call it the concentric multi-rectangle. Similarly, the multi-rectangle construction method for the polygon prediction box is the same.

\textbf{ Loss function:} The regression of the rectangular box needs to consider the degree of overlap, distance, and aspect ratio. Among the current Iou-based loss function\cite{index-38,index-39,index-40,index-41}, efficient IoU loss(EIoU) has best performance. This paper uses EIoU Loss to calculate the loss of the ground truth-predicted pairs of the above-mentioned concentric multi-rectangle. By this means, the position regression of each boundary point of the polygon can be achieved. As shown in Fig.\ref{22rect}, the procedure to calculate ground truth-predicted pairs for $Rect_i^{gt}$-$Rect_i^{pd}$ is as follows:
$$
\begin{array}{l}
EIo{U_i} = Io{U_i} - \displaystyle{\frac{{{\rho ^2}({C^{gt}},{C^{pd}})}}{{{{({c_i}\_diag)}^2}}}}\\
{\kern 1pt} {\kern 1pt} {\kern 1pt} {\kern 1pt} {\kern 1pt} {\kern 1pt} {\kern 1pt} {\kern 1pt} {\kern 1pt} {\kern 1pt} {\kern 1pt} {\kern 1pt} {\kern 1pt} {\kern 1pt} {\kern 1pt} {\kern 1pt} {\kern 1pt} {\kern 1pt} {\kern 1pt} {\kern 1pt} {\kern 1pt} {\kern 1pt} {\kern 1pt} {\kern 1pt} {\kern 1pt} {\kern 1pt} {\kern 1pt} {\kern 1pt} {\kern 1pt} {\kern 1pt} {\kern 1pt} {\kern 1pt} {\kern 1pt} {\kern 1pt} {\kern 1pt} {\kern 1pt} {\kern 1pt} {\kern 1pt} {\kern 1pt} {\kern 1pt} {\kern 1pt} {\kern 1pt} {\kern 1pt} {\kern 1pt} {\kern 1pt} {\kern 1pt} {\kern 1pt} {\kern 1pt} {\kern 1pt} {\kern 1pt} {\kern 1pt} {\kern 1pt} {\kern 1pt} {\kern 1pt} {\kern 1pt} {\kern 1pt} {\kern 1pt} {\kern 1pt} {\kern 1pt} {\kern 1pt} {\kern 1pt} {\kern 1pt} - \displaystyle{\frac{{{{(Rect_i^{gt}\_w - Rect_i^{pd}\_w)}^2}}}{{{{({c_i}\_w)}^2}}}}\\
{\kern 1pt} {\kern 1pt} {\kern 1pt} {\kern 1pt} {\kern 1pt} {\kern 1pt} {\kern 1pt} {\kern 1pt} {\kern 1pt} {\kern 1pt} {\kern 1pt} {\kern 1pt} {\kern 1pt} {\kern 1pt} {\kern 1pt} {\kern 1pt} {\kern 1pt} {\kern 1pt} {\kern 1pt} {\kern 1pt} {\kern 1pt} {\kern 1pt} {\kern 1pt} {\kern 1pt} {\kern 1pt} {\kern 1pt} {\kern 1pt} {\kern 1pt} {\kern 1pt} {\kern 1pt} {\kern 1pt} {\kern 1pt} {\kern 1pt} {\kern 1pt} {\kern 1pt} {\kern 1pt} {\kern 1pt} {\kern 1pt} {\kern 1pt} {\kern 1pt} {\kern 1pt} {\kern 1pt} {\kern 1pt} {\kern 1pt} {\kern 1pt} {\kern 1pt} {\kern 1pt} {\kern 1pt} {\kern 1pt} {\kern 1pt} {\kern 1pt} {\kern 1pt} {\kern 1pt} {\kern 1pt} {\kern 1pt} {\kern 1pt} {\kern 1pt} {\kern 1pt} {\kern 1pt} {\kern 1pt} {\kern 1pt} {\kern 1pt} - \displaystyle{\frac{{{{(Rect_i^{gt}\_h - Rect_i^{pd}\_h)}^2}}}{{{{({c_i}\_h)}^2}}}}
\end{array}  \eqno{(4)}
$$ where $i = 0,1,...,N - 3$, ${\rho ^2}({C^{gt}},{C^{pd}})$ represents the Euclidean distance between the centroid of the ground truth box and the predicted box (as well as the center of $Rect_i^{gt}$-$Rect_i^{pd}$), ${c_i}\_w$, ${c_i}\_h$ and ${c_i}\_diag$ are the width, height, and diagonal length of the circumscribe rectangle respectively of $Rect_i^{gt}$, $Rect_i^{pd}$. $Rect_i^{gt}\_w$ and $Rect_i^{gt}\_h$ are the width and height of the ground truth rectangular box. $Rect_i^{pd}\_w$ and $Rect_i^{pd}\_h$ are the width and height of the prediction rectangular box. Then EIoU loss can be computed:
$$
\begin{array}{l}
EIoU\_Los{s_i} = 1 - EIo{U_i}
\end{array} \eqno{(5)}
$$

\par Finally, we obtain $N-2$ loss terms that jointly participate in the position regression calculation.

\subsection{Adaptive Weighting Strategies of Multi-loss}
The overall loss includes multi-point regression loss, confidence loss, and classification loss. Among these losses, multi-point regression loss requires accumulation of single-point loss, significantly increasing the number of loss terms. In order to balance the different loss components, this experiment draws on the weight adjustment methods of multi-task learning. In this paper, we utilizes the DWA\cite{index-11} adjustment strategy. The weight of the $j$-th loss term is determined as follows:

$$
\begin{array}{l}
\displaystyle{\frac{{n \times \exp ({r^j}/T)}}{{\sum\limits_{i = 1}^n {\exp ({r^i}/T)} }}}
\end{array} \eqno{(6)}
$$
where $n$ is the number of loss items, $r^{i}$ is loss ratio of epoch-$i$, $T$ is temperature. There are two ways to calculate loss ratio:
\par (1)the ratio of epoch $i$ loss to epoch $i-1$ loss
$$
\begin{array}{l}
\displaystyle{r^{i}=\frac{L_{i}}{L_{i-1}}}
\end{array} \eqno{(7)}
$$
\par (2)the ratio of epoch-$i$ loss to average loss in $i-1$ epochs:
$$
\begin{array}{l}
\displaystyle{r^{i}=\frac{L_{i}}{\sum_{j=1}^{i-1} L_{j}/(i-1)}}
\end{array} \eqno{(8)}
$$
\par Comparing the calculation methods of two loss rates, the mean loss ratio is more stable. Mean loss can effectively reduce the impact of  loss fluctuations during the training process on the loss ratio. Welford algorithm\cite{index-37} is used for calculating the mean loss , and the formula is:
$$ 
\begin{array}{l}
\displaystyle{\bar{x}_{n} = \bar{x}_{n-1} + \frac{x_n - \bar{x}_{n-1}}{n}}
\end{array} \eqno{(9)}
$$where $x_{n}$ is epoch-$n$ loss in our experiment. The loss rate is calculated in the form of loss ratio, which can transform losses of different scales into the same scale. According to the definition of the loss ratio, the better the training effect of a certain regression task, the faster the loss decreases. Similarly, the smaller the loss rate, the smaller the weight assigned to it. This means the model focuses more on learning the part with a poorer training effect. This strategy meets the requirement of balancing the losses of each part. However, the temperature coefficient can also smooth the weights of each part. The higher the temperature coefficient, the more evenly the weights are distributed. When the temperature coefficient tends to infinity, the loss weight of each part is the same. Setting a temperature coefficient higher than 1 can moderately increase the weight of the part with a lower loss rate. Compared with GradNorm\cite{index-33}, there is no extra backward process required, so DWA can simplify computation. 

\section{EXPERIMENTS AND RESULTS}

This paper designs position regression loss function and multi-loss weighting strategy based on the promoted YOLOX-s\cite{index-14} detector. This section mainly introduces the process and results of the experiment.\par \textbf{Implementation details}: Woodscape\cite{index-42} is used to train model and verify our method. This experiment randomly selects about 5000 training data and about 200 test data for mAP calculation. There are 8 categories, which are: person, bicycle, car, motorcycle, bus, train, truck and traffic light. Our model is trained on two RTX3090 GPUs for 500 epochs with batch size 32. But a large number of parameters and high computational cost of multi-point representation complicate the data augmentation.Therefore, we only used simple data augmentation techniques such as color space transformation, aspect ratio distortion, translation, rotation, and flipping to give the detector basic generalization ability. The weighting strategy of loss function involves the definition of temperature and loss rates. Experiment A set the temperature coefficient as T=20, and the loss rate as the ratio of current loss to previous single loss. Experiment B adopts 22-rectangle plus poly IoU, further exploring the effect of temperature and loss ratio. 

\subsection{Multi-point IoU Loss Function}
\begin{table}[htbp]
    \centering
    \caption{The exploration for concentric rectangle}
    \label{tab:1} 
    \tabcolsep=1cm  
    \renewcommand{\arraystretch}{1.5}  
    \begin{tabular}{cccccc} 
    \hline\hline\noalign{\smallskip}
    \multirow{1}*{\textbf{Loss Function}} &  \multirow{1}*{\textbf{mAP}}\\
    \midrule
    \multirow{1}*{22 vertex-shared rectangle}  & 17.77 $\%$\\
    \midrule
    \multirow{1}*{22 center-shared rectangle}  & \textbf{25.97 $\%$} \\  
    \bottomrule
    \end{tabular}
    
\end{table}

At the beginning, we designed several different forms of multi-rectangle construction and selected a small sample size of data as the training set.\par $\bullet$ 22 vertex-shared rectangles: Our boundary points aligned with the centroid axis are grouped into two pairs. Each pair forms a rectangle with the centroid as one of the vertices (the diagonal is the line connecting the two boundary points). The remaining 20 boundary points are used to construct 20 rectangles, each with the centroid as one of the vertices (the diagonal is the line connecting the centroid and the boundary point). Each rectangle must have one vertex at the centroid.\par $\bullet$ 22 concentric rectangles: Section \uppercase\expandafter{\romannumeral3} explains this method in detail.\par Loss functions both use EIoU loss. As shown in Table \ref{tab:1}, Scheme 2 performs significantly better than another. We believe that the accuracy of the centroid has a greater impact on the final detection results compared to the multiple points.The shared centroid design in Scheme 2 precisely aggregates the power of multiple scattered rectangle regressions. Such effect enables the multiple rectangles to converge faster to the same center point, that is, the centroid. Although Scheme 1 also has the feature of shared vertices, the vertices are jointly determined by the width and height, making regression more difficult. Scheme 1 does not have any commonality between multiple rectangles, and therefore no collective force is formed.\par Based on above analysis, three types of position regression loss functions (poly IoU, 24 concentric circles loss\cite{index-11} and 22 concentric rectangles EIoU) were used either individually or in combination for contrast experiment. The mAP results are shown in Table \ref{tab:2}. Using the Poly IoU alone for position regression resulted in poor detection performance. After analysis, there are three main reasons explaining this phenomenon. Firstly, Rectangular bounding boxes can use IoU loss better than Euclidean distance loss because of the strong correlation between the four points. However, the boundary points of a polygonal box are relatively independent. Hence area loss alone cannot accurately predict multiple points.The direct analogy of Poly IoU to rectangular IoU without adaptive improvement loses effect. Secondly, Poly IoU loss struggles to achieve synchronous regression of multiple points. By splitting Poly IoU loss into 22 rectangular IoU losses, each point can be regressed separately and aggregated to form the overall regression. Thirdly, Poly IoU does not restrict the shape and orientation of the predicted bounding boxes. The same IoU value corresponds to different shapes and orientations, decreasing prediction accuracy.\par
\begin{figure*}[htpb]
\centering
\includegraphics[width=\textwidth]{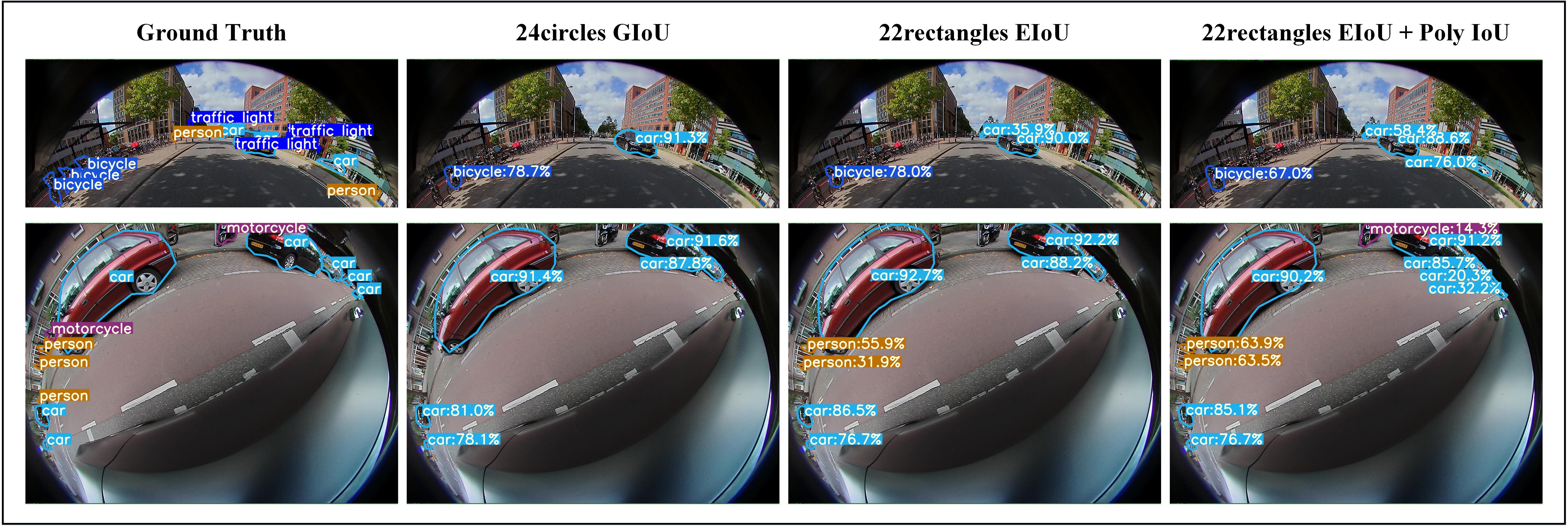}
\caption{Ground truth and detection results.}
\label{result}
\end{figure*}

 Compared to Poly IoU loss, the detection performance is significantly improved when using 24 concentric circles GIoU loss or the 22 concentric rectangles EIoU loss. If Poly IoU loss is combined with them, it can further improve the detection performance. Overall, the combination of the proposed concentric rectangles EIou loss and the Poly-IoU loss is a good choice for multi-point representation regression strategy. The detection results and ground thruth of each method are shown in Fig. \ref{result}.

\par 
\begin{table}[htbp]
    \centering
    \caption{The mAP of different IoU loss}
    \label{tab:2} 
    \tabcolsep=1cm  
    \renewcommand{\arraystretch}{1.5}  
    \begin{tabular}{cccccc} 
    \hline\hline\noalign{\smallskip}
    \multirow{1}*{\textbf{Loss Function}} &  \multirow{1}*{\textbf{mAP}}\\
    \midrule
    \multirow{1}*{Poly IoU}  & 3.12$\%$ \\
    \midrule
    \multirow{1}*{24 circles GIoU}  & 32.24$\%$ \\
    \midrule
    \multirow{1}*{22 rectangles EIoU}  & 34.10$\%$ \\
    \midrule
    \multirow{1}*{24 circles GIoU + Poly IoU}  & 34.28$\%$ \\
            
    \midrule
    \multirow{1}*{22 rectangles EIoU + Poly IoU}  & \textbf{36.09$\%$}\\
               
    \bottomrule
    \end{tabular}
    
\end{table}
\subsection{Exploration about Loss Weight}

\par \textbf{Exploration about temperature}: At the initial stage of the experiment, the temperature coefficient is set to 1. However, it  is found that the losses of each part fluctuat greatly, and the final prediction performance is poor. Therefore, the temperature coefficient is adjusted, resulting in a change in the weight distribution. Through a series of comparisons of temperature coefficients, the experiment find that setting the temperature coefficient to a value greater than 1 can reduce the instability of the loss and improve the prediction performance. Since the temperature coefficient has a smoothing effect on the weights, increasing the temperature coefficient can moderately reduce the difference between the weights. This can avoid sharp changes in weights between different batches. Meanwhile, increasing the temperature can moderately raise the weight of the loss with good training performance. This trend can reduce the risk of falling into a local minimum to some extent. But too high temperature may cause the weights to lose significant discrimination. Thus the model cannot focus on learning the poorer performing regressions. Therefore, it is necessary to find the optimal value through experimental comparison. As shown in Table\ref{tab:3}, based on the mAP under different temperature coefficients, the final temperature is set to 20.
\begin{table}[htbp]
    \centering
    \caption{The mAP of different temperature}
    \label{tab:3} 
    \tabcolsep=1cm  
    \renewcommand{\arraystretch}{1.5}  
    \begin{tabular}{cccccc} 
    \hline\hline\noalign{\smallskip}
    \multirow{1}*{\textbf{Temperature}} &  \multirow{1}*{\textbf{mAP}}\\
    \midrule
    \multirow{1}*{T=1}  & 30.95$\%$ \\
    \midrule
    \multirow{1}*{T=20}  & \textbf{40.60$\%$} \\
    \midrule
    \multirow{1}*{T=50}  & 35.55$\%$ \\  
    \bottomrule
    \end{tabular}
    
\end{table}

\begin{figure}[htpb]
\centering
\includegraphics[width=0.7\textwidth]{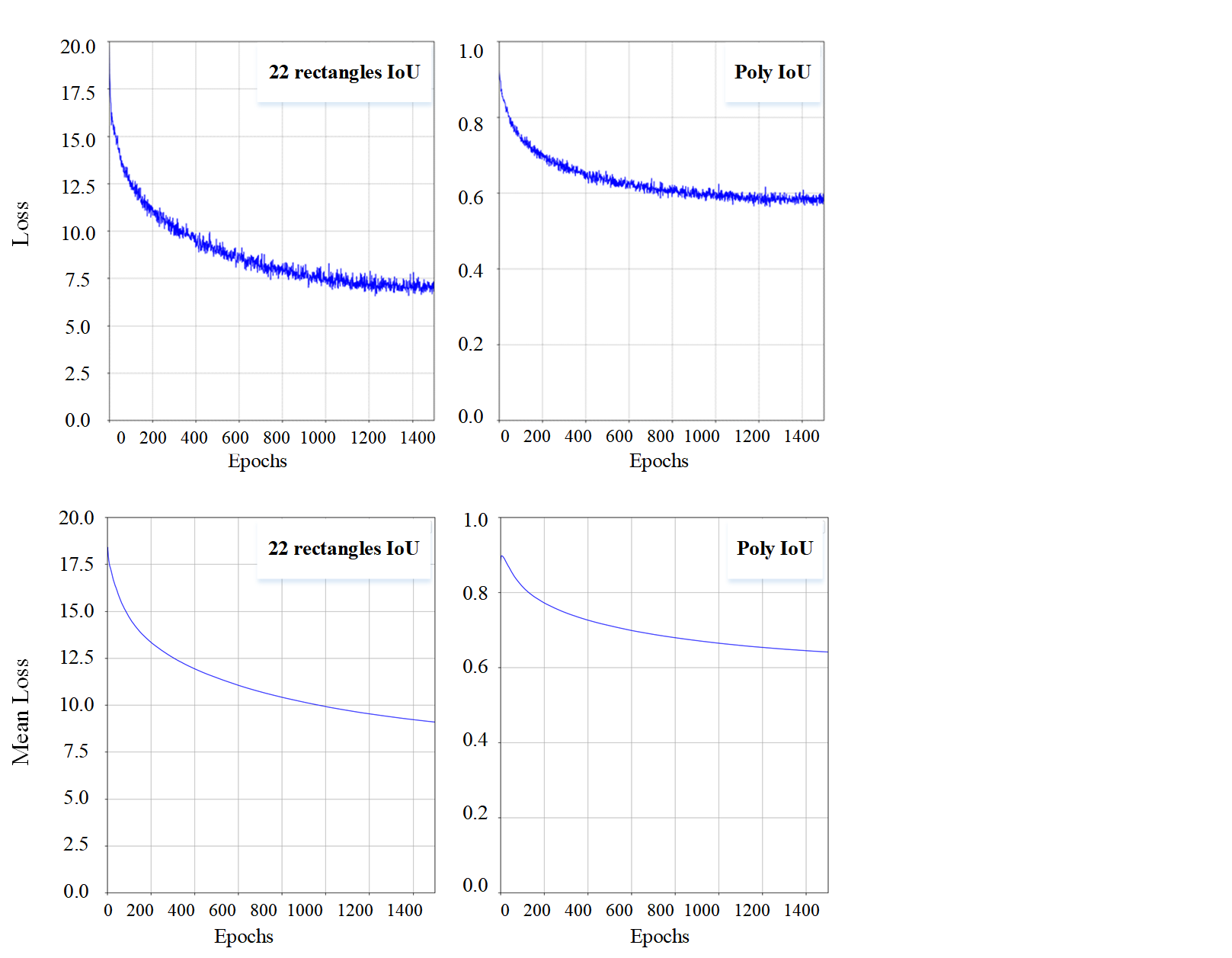}
\caption{Comparison of mean loss until the last epoch and 
current loss.}
\label{losses line chart}
\end{figure}

\textbf{Exploration about loss ratio:} The line chart of losses is shown in Fig.\ref{losses line chart}. It can be observed that the single-epoch loss fluctuates greatly, but the average loss value is relatively smooth. If the loss ratio is defined by single-epoch loss, the value strongly depends on the loss in the previous epoch. Then the irregular fluctuations in the single-epoch loss will affect the accuracy of the loss ratio and weight allocation. Therefore, the experiment improved the above loss ratio by changing the single loss to a more stable average loss, making the calculation of the loss rate more accurate.
\section{CONCLUSIONS AND FUTURE WORK}
In this paper, we studied regression loss for multi-point representation and designed delicate comparative experiment.
To sum up, we draw following conclusions.
\par(1) It is difficult to acquire compatible results only by Poly IoU loss. To solve this problem, we split multi-point group to single points, and design the corresponding loss items for each point. The experiment shows that the items related to dense distance can significantly improve detection performance. Moreover, the combination of Poly IoU Loss and Concentric Rectangles Loss has better results.

\par (2) Increasing the temperature moderately can decrease the discrimination of weights and enhance the stability of total loss. In addition, mean loss can help reducing the impact of loss fluctuations and make weight allocation more effective.

Our future work is oriented to improve generalization of model and explore a mechanism for selecting points according to object size and shape. We believe that multi-point representation will gain persistent attention.

\section*{ACKNOWLEDGMENT}
This work was partly supported by National Natural Science Foundation of China (Grant No. U1913203, 61973034, 62233002 and CJSP Q2018229). The authors would like to thank Tianji Jiang, Jiadong Tang, Zhaoxiang Liang, Dianyi Yang, and all other members of ININ Lab of Beijing Institute of Technology for their contribution to this work.

\bibliographystyle{ieeetr}
\bibliography{reference}

\begin{thebibliography}{10}

\bibitem{index-2}
J.~Ma, W.~Shao, H.~Ye, L.~Wang, H.~Wang, Y.~Zheng, and X.~Xue,
  ``Arbitrary-oriented scene text detection via rotation proposals,'' {\em IEEE
  transactions on multimedia}, vol.~20, no.~11, pp.~3111--3122, 2018.

\bibitem{index-3}
M.~Liao, B.~Shi, and X.~Bai, ``Textboxes++: A single-shot oriented scene text
  detector,'' {\em IEEE transactions on image processing}, vol.~27, no.~8,
  pp.~3676--3690, 2018.

\bibitem{index-5}
H.~Yang, R.~Deng, Y.~Lu, Z.~Zhu, Y.~Chen, J.~T. Roland, L.~Lu, B.~A. Landman,
  A.~B. Fogo, and Y.~Huo, ``Circlenet: Anchor-free detection with circle
  representation,'' {\em arXiv preprint arXiv:2006.02474}, 2020.

\bibitem{index-4}
W.~Dong, P.~Roy, C.~Peng, and V.~Isler, ``Ellipse r-cnn: Learning to infer
  elliptical object from clustering and occlusion,'' {\em IEEE Transactions on
  Image Processing}, vol.~30, pp.~2193--2206, 2021.

\bibitem{index-7}
T.~Li, G.~Tong, H.~Tang, B.~Li, and B.~Chen, ``Fisheyedet: A self-study and
  contour-based object detector in fisheye images,'' {\em IEEE Access}, vol.~8,
  pp.~71739--71751, 2020.

\bibitem{index-8}
H.~Rashed, E.~Mohamed, G.~Sistu, V.~R. Kumar, C.~Eising, A.~El-Sallab, and
  S.~Yogamani, ``Fisheyeyolo: Object detection on fisheye cameras for
  autonomous driving,'' in {\em Machine Learning for Autonomous Driving NeurIPS
  2020 Virtual Workshop}, vol.~8, 2020.

\bibitem{index-9}
H.~Rashed, E.~Mohamed, G.~Sistu, V.~R. Kumar, C.~Eising, A.~El-Sallab, and
  S.~Yogamani, ``Generalized object detection on fisheye cameras for autonomous
  driving: Dataset, representations and baseline,'' in {\em Proceedings of the
  IEEE/CVF Winter Conference on Applications of Computer Vision},
  pp.~2272--2280, 2021.

\bibitem{index-10}
E.~Xie, P.~Sun, X.~Song, W.~Wang, X.~Liu, D.~Liang, C.~Shen, and P.~Luo,
  ``Polarmask: Single shot instance segmentation with polar representation,''
  in {\em Proceedings of the IEEE/CVF conference on computer vision and pattern
  recognition}, pp.~12193--12202, 2020.

\bibitem{index-11}
X.~Xu, Y.~Gao, H.~Liang, Y.~Yang, and M.~Fu, ``Fisheye object detection based
  on standard image datasets with 24-points regression strategy,'' in {\em 2022
  IEEE/RSJ International Conference on Intelligent Robots and Systems (IROS)},
  pp.~9911--9918, IEEE, 2022.

\bibitem{index-12}
S.~Liu, E.~Johns, and A.~J. Davison, ``End-to-end multi-task learning with
  attention,'' in {\em Proceedings of the IEEE/CVF conference on computer
  vision and pattern recognition}, pp.~1871--1880, 2019.

\bibitem{index-13}
R.~Girshick, ``Fast r-cnn,'' in {\em Proceedings of the IEEE international
  conference on computer vision}, pp.~1440--1448, 2015.

\bibitem{index-14}
Z.~Ge, S.~Liu, F.~Wang, Z.~Li, and J.~Sun, ``Yolox: Exceeding yolo series in
  2021,'' {\em arXiv preprint arXiv:2107.08430}, 2021.

\bibitem{index-16}
C.-Y. Wang, A.~Bochkovskiy, and H.-Y.~M. Liao, ``Yolov7: Trainable
  bag-of-freebies sets new state-of-the-art for real-time object detectors,''
  {\em arXiv preprint arXiv:2207.02696}, 2022.

\bibitem{index-17}
Z.~Duan, O.~Tezcan, H.~Nakamura, P.~Ishwar, and J.~Konrad, ``Rapid:
  rotation-aware people detection in overhead fisheye images,'' in {\em
  Proceedings of the IEEE/CVF Conference on Computer Vision and Pattern
  Recognition Workshops}, pp.~636--637, 2020.

\bibitem{index-18}
O.~Krams and N.~Kiryati, ``People detection in top-view fisheye imaging,'' in
  {\em 2017 14th IEEE international conference on advanced video and signal
  based surveillance (AVSS)}, pp.~1--6, IEEE, 2017.

\bibitem{index-20}
B.~Arsenali, P.~Viswanath, and J.~Novosel, ``Rotinvmtl: Rotation invariant
  multinet on fisheye images for autonomous driving applications,'' in {\em
  Proceedings of the IEEE/CVF International Conference on Computer Vision
  Workshops}, pp.~0--0, 2019.

\bibitem{index-21}
E.~H. Nguyen, H.~Yang, R.~Deng, Y.~Lu, Z.~Zhu, J.~T. Roland, L.~Lu, B.~A.
  Landman, A.~B. Fogo, and Y.~Huo, ``Circle representation for medical object
  detection,'' {\em IEEE transactions on medical imaging}, vol.~41, no.~3,
  pp.~746--754, 2021.

\bibitem{index-22}
M.~Cao, S.~Ikehata, and K.~Aizawa, ``Field-of-view iou for object detection in
  360 $\{$$\backslash$deg$\}$ images,'' {\em arXiv preprint arXiv:2202.03176},
  2022.

\bibitem{index-23}
P.~Zhao, A.~You, Y.~Zhang, J.~Liu, K.~Bian, and Y.~Tong, ``Spherical criteria
  for fast and accurate 360 object detection,'' in {\em Proceedings of the AAAI
  Conference on Artificial Intelligence}, vol.~34, pp.~12959--12966, 2020.

\bibitem{index-24}
R.~Caruana, {\em Multitask learning}.
\newblock Springer, 1998.

\bibitem{index-25}
O.~Sener and V.~Koltun, ``Multi-task learning as multi-objective
  optimization,'' {\em Advances in neural information processing systems},
  vol.~31, 2018.

\bibitem{index-26}
X.~Lin, H.~Baweja, G.~Kantor, and D.~Held, ``Adaptive auxiliary task weighting
  for reinforcement learning,'' {\em Advances in neural information processing
  systems}, vol.~32, 2019.

\bibitem{index-27}
Y.~Du, W.~M. Czarnecki, S.~M. Jayakumar, M.~Farajtabar, R.~Pascanu, and
  B.~Lakshminarayanan, ``Adapting auxiliary losses using gradient similarity,''
  {\em arXiv preprint arXiv:1812.02224}, 2018.

\bibitem{index-29}
A.~Kendall, Y.~Gal, and R.~Cipolla, ``Multi-task learning using uncertainty to
  weigh losses for scene geometry and semantics,'' in {\em Proceedings of the
  IEEE conference on computer vision and pattern recognition}, pp.~7482--7491,
  2018.

\bibitem{index-30}
A.~Kendall and Y.~Gal, ``What uncertainties do we need in bayesian deep
  learning for computer vision?,'' {\em Advances in neural information
  processing systems}, vol.~30, 2017.

\bibitem{index-31}
F.~Zheng, C.~Deng, X.~Sun, X.~Jiang, X.~Guo, Z.~Yu, F.~Huang, and R.~Ji,
  ``Pyramidal person re-identification via multi-loss dynamic training,'' in
  {\em Proceedings of the IEEE/CVF conference on computer vision and pattern
  recognition}, pp.~8514--8522, 2019.

\bibitem{index-32}
S.~Liu, Y.~Liang, and A.~Gitter, ``Loss-balanced task weighting to reduce
  negative transfer in multi-task learning,'' in {\em Proceedings of the AAAI
  conference on artificial intelligence}, vol.~33, pp.~9977--9978, 2019.

\bibitem{index-33}
Z.~Chen, V.~Badrinarayanan, C.-Y. Lee, and A.~Rabinovich, ``Gradnorm: Gradient
  normalization for adaptive loss balancing in deep multitask networks,'' in
  {\em International conference on machine learning}, pp.~794--803, PMLR, 2018.

\bibitem{index-34}
M.~Crawshaw and J.~Ko{\v{s}}eck{\'a}, ``Slaw: Scaled loss approximate weighting
  for efficient multi-task learning,'' {\em arXiv preprint arXiv:2109.08218},
  2021.

\bibitem{index-35}
S.~Verboven, M.~H. Chaudhary, J.~Berrevoets, V.~Ginis, and W.~Verbeke,
  ``Hydalearn: Highly dynamic task weighting for multitask learning with
  auxiliary tasks,'' {\em Applied Intelligence}, pp.~1--15, 2022.

\bibitem{index-36}
R.~Groenendijk, S.~Karaoglu, T.~Gevers, and T.~Mensink, ``Multi-loss weighting
  with coefficient of variations,'' in {\em Proceedings of the IEEE/CVF winter
  conference on applications of computer vision}, pp.~1469--1478, 2021.

\bibitem{index-37}
B.~Welford, ``Note on a method for calculating corrected sums of squares and
  products,'' {\em Technometrics}, vol.~4, no.~3, pp.~419--420, 1962.

\bibitem{index-39}
Y.-F. Zhang, W.~Ren, Z.~Zhang, Z.~Jia, L.~Wang, and T.~Tan, ``Focal and
  efficient iou loss for accurate bounding box regression,'' {\em
  Neurocomputing}, vol.~506, pp.~146--157, 2022.

\bibitem{index-38}
Z.~Zheng, P.~Wang, W.~Liu, J.~Li, R.~Ye, and D.~Ren, ``Distance-iou loss:
  Faster and better learning for bounding box regression,'' in {\em Proceedings
  of the AAAI conference on artificial intelligence}, vol.~34,
  pp.~12993--13000, 2020.

\bibitem{index-40}
J.~Yu, Y.~Jiang, Z.~Wang, Z.~Cao, and T.~Huang, ``Unitbox: An advanced object
  detection network,'' in {\em Proceedings of the 24th ACM international
  conference on Multimedia}, pp.~516--520, 2016.

\bibitem{index-41}
H.~Rezatofighi, N.~Tsoi, J.~Gwak, A.~Sadeghian, I.~Reid, and S.~Savarese,
  ``Generalized intersection over union: A metric and a loss for bounding box
  regression,'' in {\em Proceedings of the IEEE/CVF conference on computer
  vision and pattern recognition}, pp.~658--666, 2019.

\bibitem{index-42}
S.~Yogamani, C.~Hughes, J.~Horgan, G.~Sistu, P.~Varley, D.~O'Dea,
  M.~Uric{\'a}r, S.~Milz, M.~Simon, K.~Amende, {\em et~al.}, ``Woodscape: A
  multi-task, multi-camera fisheye dataset for autonomous driving,'' in {\em
  Proceedings of the IEEE/CVF International Conference on Computer Vision},
  pp.~9308--9318, 2019.

\end{thebibliography}
\end{document}